# CNN-transformer mixed model for object detection

Student Name:Wenshuo Li
Student Number: 210276415
Supervisor Name: Ioannis Patras
Programme of study: Machine Learning
for Visual Data Analytics

*Abstract*—Object detection, one of the three main tasks of computer vision, has been used in various applications. The main process is to use deep neural networks to extract the features of an image and then use the features to identify the class and location of an object. Therefore, the main direction to improve the accuracy of object detection tasks is to improve the neural network to extract features better. In this paper, I propose a convolutional module with a transformer[1], which aims to improve the recognition accuracy of the model by fusing the detailed features extracted by CNN[2] with the global features extracted by a transformer and significantly reduce the computational effort of the transformer module by deflating the feature mAP. The main execution steps are convolutional downsampling to reduce the feature map size, then self-attention calculation and upsampling, and finally concatenation with the initial input. In the experimental part, after splicing the block to the end of YOLOv5n[3] and training 300 epochs on the coco dataset, the mAP improved by 1.7% compared with the previous YOLOv5n, and the mAP curve did not show any saturation phenomenon, so there is still potential for improvement. After 100 rounds of training on the Pascal VOC dataset, the accuracy of the results reached 81%, which is 4.6 better than the faster RCNN[4] using resnet101[5] as the backbone, but the number of parameters is less than one-twentieth of it.

*Keywords*—CNN, transformer, computer vision, object detection.

## I. INTRODUCTION

### A. Computer vision

Machine learning refers to the process of extracting knowledge from data by computer, deep learning is a branch of machine learning, which uses a multi-layer neural network structure to learn data, deep learning often has a large number of parameters and computational amount. With the increase in computer computing speed, the application of deep learning is becoming more and more widespread, among which computer vision is the main application area of deep learning.

Computer vision is a subset of artificial intelligence, which generally uses machine learning or deep learning to analyze images or videos for different purposes. Computer vision has a wide range of applications, such as image classification, face recognition, driving assistance, industrial product inspection, etc. The three classical tasks of computer vision are image classification, object detection, and image segmentation. Computer vision tasks mainly use deep learning such as CNN or vision transformer to extract image features and then use these features for different tasks.

### B. Object detection

Object detection means inputting a picture or video, identifying the object category, and displaying the size and position of each object and the confidence value using a rectangular box.

Initially, object detection methods mainly used human-defined image features and then used machine learning to detect them, such as the bag-of-words [6] method. RCNN[7] started to use deep learning methods to extract features from candidate boxes for classification using neural networks, which significantly improved the accuracy. In deep learning-based target detection methods, generally divided into one-stage methods and two-stage methods, two-stage methods, such as faster RCNN[4], first generates candidate boxes by a neural network module (RPN) for distinguishing objects from the background. Then the candidate boxes generated in the previous step are classified and resized in position.

The one-stage method, such as YOLO[8], directly uses a neural network to classify and resize the position of the bounding box. Therefore, the one-stage method is fast, and its accuracy depends on the extraction of image features by the neural network. Therefore, how to better extract the features of the image using deep learning networks becomes the main goal to improve the performance of object detection.

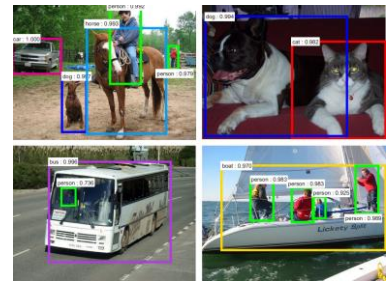

Figure 1: The final output of target detection is a class of objects and a bounding box with location information

The main deep learning network used to extract image features is CNN, which extracts local features by sharing weights through a sliding convolutional kernel. The shared weights make it much less computationally complex than traditional fully-connected neural networks and allow it to extract similar features for similar objects at different locations, which is in line with the characteristics of images, and the localities of images tend to be more closely related to each other. Because of these image properties, convolutional neural networks have a good performance and have been in the dominant position. Convolutional neural networks achieve feature extraction from small localities to large localities by stacking multiple layers of convolution and pooling blocks.

CNN is no longer the mainstream after the introduction of the vision transformer. The transformer is a deep learning network that computes the relationship between two input vectors[9], because it computes two pairs, it can extract global features at the beginning compared with a convolutional neural network which extracts local features, but its computational complexity is squared, so it is very computationally intensive.

*C. Objectives of the project*

The goal of this project is to improve the architecture of the neural network for the object detection task so that it can extract features better and thus improve the accuracy of detection. In this project, I propose a hybrid module of convolutional neural network and transformer and use it for the object detection task. It can use a transformer to extract global features but with relatively low computational cost. In this block, the feature map is downsampled through a convolutional layer before it enters the transformer computation, in order to reduce the size of the feature map in order to significantly reduce the computational volume of the transformer. This residual structure allows the features extracted by the transformer to be fused with the previous features and ensures that the learning effect is no worse than the previous one.

After applying the block to the end of YOLOv5n and training 300 epochs on the Coco dataset, the mAP was improved by 1.7% compared to the previous YOLOv5n, and the mAP curve did not saturate, so there is still potential for improvement. After 100 rounds of training on the Pascal VOC dataset, the accuracy reached 81%, which is 4.6% higher than the faster RCNN using resnet101 as the backbone, but the number of participants is less than one-twentieth of it.

## II. RELATED WORK

*A. RCNN*

RCNN[7] is the first deep learning model for object detection, which is a milestone. The process of RCNN is divided into the following steps: 1 it first uses selective search algorithm to merge different regions of the image, and finally generates about two thousand candidate boxes, 2 then the regions in the candidate boxes are fed into the trained neural network Alexnet to extract features, 3 then the extracted features are classified using SVM[10], and the class of the object in the candidate box is output. 4 After removing the duplicate candidate frames using non-maximum suppression, the positions of the candidate frames are modified using regression.

The problem with RCNN is that the selective search algorithm is computationally expensive and the model is complex to train, requiring the training of deep neural networks, SVMs, and regressors. Therefore, the subsequent Fast RCNN combines Alexnet[11], SVM, and regression into a single network for feature extraction, classification, and correction of the bounding box position. The subsequent Faster RCNN also uses neural networks instead of selective search algorithm to generate candidate boxes, which significantly reduces the computational amount and increases the speed.

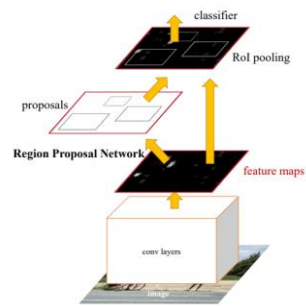

Figure 2: The two stages of faster RCNN

*B. YOLO*

Compared to Faster RCNN, which is a two-step method to generate candidate boxes and then predict them, YOLO[8] directly uses a network to predict and adjust the bounding boxes. It splits the input image into multiple grids, each grid point represents the center of an object and predicts multiple bounding boxes including its position and the class of the object inside. Each bounding box has a confidence value, which represents the probability that the bounding box contains an object, so that candidate boxes with a small probability of containing objects can be removed by non-maximal suppression. The initial shortcoming of YOLO was that it was not effective in detecting small objects and new size objects. To solve these problems, in subsequent versions, better backbone networks such as DarkNet19[12] and DarkNet53, CSPDarkNet53[13], new modules and structures incorporating multi-scale features such as SPPF[14], PAN[15], and improved loss calculation and use of image enhancement were proposed.

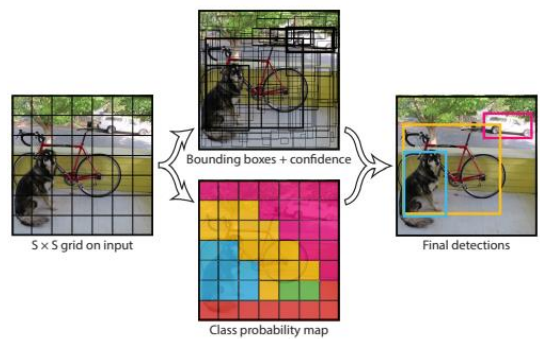

Figure 3: Each grid in YOLO predicts the size and class and confidence of multiple bounding boxes, and finally only the bounding box with high confidence is retained.

## III. METHOD

The networks used for extracting image features in computer vision tasks, such as resnet50[16], darknet53, etc., are often formed by combining blocks with different functions to eventually form deep neural networks. Different blocks have different roles, and different combinations of structures can be adapted to different computer vision tasks.

*A. CNN+Batch normalization+Silu*

The most commonly used blocks in convolutional neural networks are convolution, batch normalization[17], activation function, and pooling block. If fully connected layers are used to extract image features, the image must be flattened into vectors, which makes it difficult to train with a large number of parameters and loses the spatial information of the image. Also, the features extracted from the same object in different places will be different. In contrast to the fully connected layer, the convolutional layer uses multiple sliding convolutional kernels as weights to compute the results of the previous layer. The convolutional operation has the feature of weight sharing, which not only makes similar objects have similar features but also significantly reduces the computational cost and facilitates the extraction of image features with spatial characteristics.

BN[17] refers to batch normalization, where the output is normalized to be concentrated around 0. Batch normalization is usually used before the activation function because the activation function is usually nonlinear and the gradient will be small if the input is too large. Batch normalization can avoid gradient disappearance and speed up the convergence during training.

Activation functions are generally nonlinear functions, such as the Sigmoid function, Relu function, Silu function, etc. The purpose of the activation function is to add nonlinear factors to the neural network to solve the nonlinear problem. If the activation function is not added, then the neural network will have only linear operations and will not be able to fit complex problems. Silu's formula is

$$silu(x) = x * sigmoid(x) = x \frac{1}{1+e^{-x}}$$

Its image is very similar to Relu[18] but with negative values, because it has a straight line on the right side, so it can effectively avoid the gradient disappearance. But it is more nonlinear than Relu, which makes it more generalizable and normalizable.

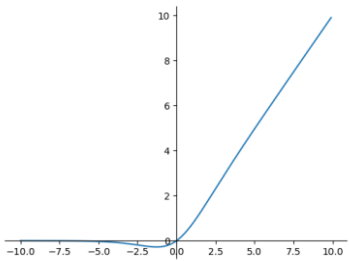

Figure 4: SILU activation function

Pooling reduces the length and width of the feature map so that the next convolution has a larger field of view. In a convolutional neural network, the size of the feature map is reduced by pooling, so that the later convolutional layers have a larger and larger field of view, and more global features can be learned. This reduction in size can be replaced by a convolutional operation with a stride greater than 1. Therefore, in this method, the convolution module is a combination of convolution operation + BN + Silu.

*B. C3 block*

C3 block is derived from the BottleNetCSP[19] architecture. C3 block reduces a convolution operation of BottleNetCSP. It concats the result of convolution with the previous one, and repeats it several times, and then convolves the result again. This allows the fusion of multiple layers of features and more detailed extraction of the target texture.

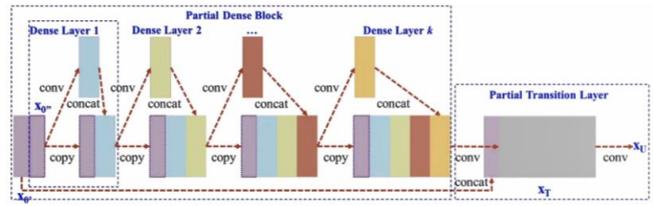

Figure 5: BottleNetCSP architecture

*C. SPPF and PAN*

Both SPP[14] and PAN[15] are a pyramid structure, they are designed to fuse features at different scales, because the fusion of features at multiple scales is of great importance for target detection. The shallow feature map has a small field of view and can only capture a small range of features, but often has more detail, so the use of shallow feature maps can help identify small objects. Deeper feature maps have a larger field of view and can capture global information, which is suitable for large objects. For object detection, the multi-scale feature fusion can improve the accuracy significantly since it can observe both global features and texture details of the object. SPPF takes the final output of the main network through several parallel pooling to obtain features of different scales and then merges them. The PAN structure is a fusion of feature maps of different depths of the backbone to obtain features of different scales and use them for final classification.

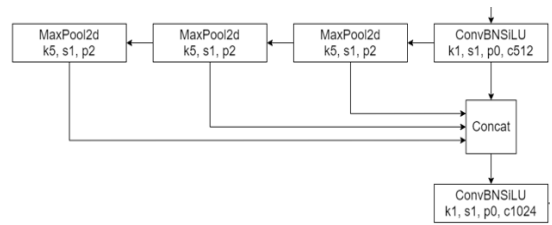

Figure 6: SPPF Pyramid Structure

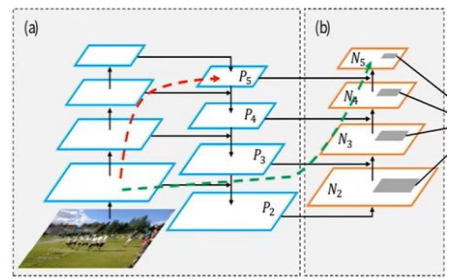

Figure 7: PAN Pyramid Structure

*D. Conv-transformer block*

The main role of the Transformer is the calculation of self-attention [20], which is a method that computes all inputs in two pairs. It generates the corresponding q,k,v vectors for each input, calculates the q of each input with the k of all other inputs to obtain the correlation α, and then multiplies α softmax with the corresponding v. The advantage of this computation is that it can compute the relationship between every two inputs, which is global in nature. In contrast, CNN uses a sliding convolution kernel, so it can only observe local information and the convolution operation cannot establish the relationship between more distant pixels. However, the transformer's two-by-two calculation method has a squared complexity, and when the input reaches a certain number, the computation becomes too large to compute.

To reduce the computation, the vision transformer[1] divides the image into 16*16 patches, if the input is a 640*640 size image, then the size of each patch is 40*40, which is obviously too large, although the global information is obtained, but not conducive to capture the details of the image.

So in order to reduce the computational cost and get the global information without losing the details of the image, I propose a block with a combination of convolution and transformer. My core idea is to fuse the detailed features extracted by the convolution operation with the global features extracted by the transformer. For example, in yolov5, there are 20*20, 40*40, and 80*80 feature maps. If the 40*40 and 80*80 feature maps are input into the transformer, the computation will be too large to compute. So I use a downsampling convolution block (Conv+BN+Silu) to reduce the size of the feature map to 20*20 and then input it into the transformer block. In the Transformer block structure, the two-dimensional feature map is first flattened into a vector and then a learnable position code is embedded, and then the multi-head attention calculation is repeated several times. In order to reduce the computational cost, I only perform the multi-head attention calculation once and set the number of multi-head to 4 (12 in the vision transformer). The next step is to revert to the original size using upsampling. After the calculation, the upsampling block is used to scale it back to its original size and then concatenate it with the initial input. Finally, the result is fused using a convolution block.

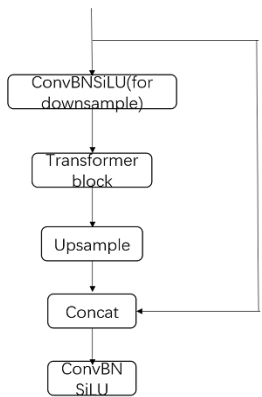

Figure 8: Conv-transformer structure

For the simplicity of the model, I added the Conv-Transformer block to the end of the YOLOv5n network, and the improved overall structure is shown in Fig9.

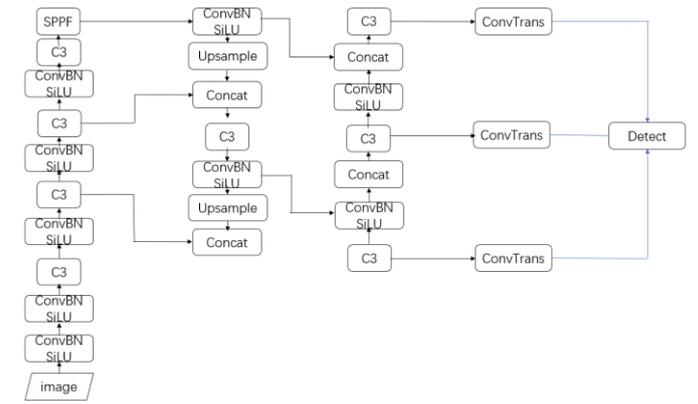

Figure 9: Overall model structure after adding Conv-transformer blocks

## IV. EXPERIMENT

### A. Experimental data set

The images in the COCO dataset are obtained from everyday scenes. For the target detection task, each object in the images has a corresponding category and bounding box. There are 80 categories of objects in the COCO dataset, which are mainly common objects in life, such as humans, vehicles, animals, household items, etc.

```
                from  n    params  module                                  arguments
  0               -1  1      1760  models.common.Conv                      [3, 16, 6, 2, 2]
  1               -1  1      4672  models.common.Conv                      [16, 32, 3, 2]
  2               -1  1      4800  models.common.C3                        [32, 32, 1]
  3               -1  1     18560  models.common.Conv                      [32, 64, 3, 2]
  4               -1  2     29184  models.common.C3                        [64, 64, 2]
  5               -1  1     73984  models.common.Conv                      [64, 128, 3, 2]
  6               -1  3    156928  models.common.C3                        [128, 128, 3]
  7               -1  1    295424  models.common.Conv                      [128, 256, 3, 2]
  8               -1  1    296448  models.common.C3                        [256, 256, 1]
  9               -1  1    164608  models.common.SPPF                      [256, 256, 5]
 10               -1  1     33024  models.common.Conv                      [256, 128, 1, 1]
 11               -1  1         0  torch.nn.modules.upsampling.Upsample    [None, 2, 'nearest']
 12          [-1, 6]  1         0  models.common.Concat                    [1]
 13               -1  1     90880  models.common.C3                        [256, 128, 1, False]
 14               -1  1      8320  models.common.Conv                      [128, 64, 1, 1]
 15               -1  1         0  torch.nn.modules.upsampling.Upsample    [None, 2, 'nearest']
 16          [-1, 4]  1         0  models.common.Concat                    [1]
 17               -1  1     22912  models.common.C3                        [128, 64, 1, False]
 18               -1  1     36992  models.common.Conv                      [64, 64, 3, 2]
 19         [-1, 14]  1         0  models.common.Concat                    [1]
 20               -1  1     74496  models.common.C3                        [128, 128, 1, False]
 21               -1  1    147712  models.common.Conv                      [128, 128, 3, 2]
 22         [-1, 10]  1         0  models.common.Concat                    [1]
 23               -1  1    296448  models.common.C3                        [256, 256, 1, False]
 24               17  1    115264  models.common.ConvTransformerBlock      [64, 64, 4]
 25               20  1    263296  models.common.ConvTransformerBlock      [128, 128, 2]
 26               23  1    854272  models.common.ConvTransformerBlock      [256, 256, 1]
 27     [24, 25, 26]  1    115005  models.yolo.Detect                      [80, [[10, 13, 16, 30,
Model summary: 333 layers, 3104989 parameters, 3104989 gradients, 5.5 GFLOPs
```

Figure 10: Detailed parameter configuration of each module in Pytorch

COCO128 is taken from the first 128 images in COCO, and is generally used to test whether the model can work properly.

Pascal VOC is a more lightweight dataset than COCO. The training set has 16551 images and the val set has 4952 images. There are 20 categories of objects in the images, mainly vehicles, households, animals and people.

### B. Image Enhancement

Image enhancement methods include:

1) hue, saturation, and exposure transformations.

2) Rotation, scaling, translation, non-vertical projection, the perspective transformation of images using a transformation matrix

3) flip up and down, flip left and right.

4) mosaic four pictures stitching: after initializing the background and selecting a center point, four pictures will be stitched to the background according to this center point. Combining multiple images can expand the diversity of objects in the picture.

5) image fusion: two images are superimposed on each other using different transparency.

6) Copy-paste: the objects in the image are divided and cropped out, and then pasted to another image where they do not overlap with the objects. For objects with a small number of samples, this method can improve the accuracy of the model to recognize them in different backgrounds

*C. Loss function*

The loss of the model consists of three components, which are the category loss of the objects in the bounding box, the confidence loss of whether the objects in the bounding box are objects or not, and the position loss of the bounding box. The object category loss is calculated using BCE loss, which is calculated for positive samples only. BCE loss is also used for confidence, but it is calculated for all samples because it is desired that the confidence value of negative samples is low so that the bounding box of negative samples can be removed by non-maximum suppression. BCE loss is calculated for positive samples only because negative samples do not have the groundtruth of the bounding box, and it uses CIoU loss. The overall formula is as follows

$$\mathcal{L}_{v5}(t_p, t_{gt}) = \sum_{k=0}^{K}\left[\alpha_k^{balance}\alpha_{box}\sum_{i=0}^{S^2}\sum_{j=0}^{B}\mathbb{1}_{kij}^{obj}\mathcal{L}_{CIoU} + \alpha_{obj}\sum_{i=0}^{S^2}\sum_{j=0}^{B}\mathbb{1}_{kij}^{obj}\mathcal{L}_{obj} + \alpha_{cls}\sum_{i=0}^{S^2}\sum_{j=0}^{B}\mathbb{1}_{kij}^{obj}\mathcal{L}_{cls}\right]$$

At the same time, the coefficients of the three scales of loss are balanced, because small targets are the most difficult to be detected, so the weight of loss for detecting small targets is increased, and the weight of large targets is reduced because they are the easiest to detect.

*D. Training*

After the model was built, since my model was added at the end of YOLOv5n with my proposed Conv-Transformer block, the previous backbone network was still YOLOv5n, so I could import the official YOLO pre-training weights. I first use coco128 to test whether the model is usable and then use the COCO dataset for formal training.

The number of training epochs is set to 300, which is the same as the official number of training epochs, for easy comparison. Batch-size parameter is set to -1, so that the maximum batch-size can be calculated according to the current device. With the Tesla p-100 device, the maximum available batch-size is calculated to be 109. The image size is set to 640*640. If the image size is not 640*640, the padding method is usually used to make up the 640*640 size. If the input image is a rectangular shape, padding will cause it to have a wide black border, which will not only affect the recognition result but also increase the computational amount.

Rectangular training means to pad the image into a size suitable for gpu calculation, such as 640*520, not necessarily strictly padding into 640*640, so as to reduce the black edge, can accelerate the training speed, improve the training accuracy, but in order to compare with the previous model, I did not enable rectangular training. Because yolov5n does not enable rectangular training. Autoanchor can generate three scales of anchors based on dataset clustering, which I did not enable to reduce computation. The optimizer uses GSD (stochastic gradient descent) and does not freeze the parameters of the backbone.

The initial learning rate and the final OneCycleLR learning rate are both chosen to be 0.01, and the GSD momentum is set to be 0.937. After adding the momentum, the stochastic gradient descent direction will take into account the historical changes to make the descent route more smooth and rely on inertia to jump out of the local optimal solution. The weight decay coefficient is set to 0.0005.

I first trained 300 epochs on the coco dataset, and each epoch took about 20 min with Tesla-p100. The loss steadily decreases and the accuracy steadily increases, and the 300 epochs are trained but still not saturated, so there is still space for improvement. Next, I saved the model and trained it for 999 epochs using Pascal Voc dataset, and the accuracy increased quickly and stopped training after 100 epochs.

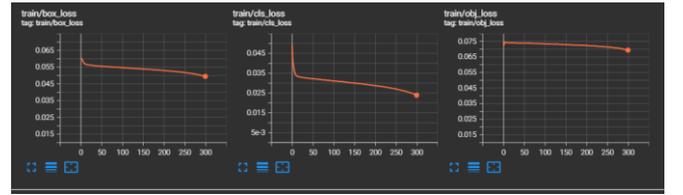

Figure 11: Training 300 epochs on COCO dataset, the loss decreases steadily and does not converge.

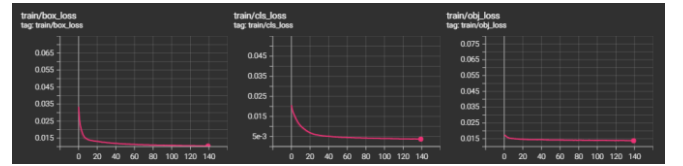

Figure 12: The loss decreases rapidly on the VOC dataset and converges around the 100th epoch.

V. RESULTS AND DISCUSSION

After training 300 epochs on the COCO dataset, the accuracy reached 47.4, a 1.7 percent improvement compared to yolov5n before the improvement, but it did not saturate, and the mAP was still on an upward trend, so if I retrained more epochs, I would get a higher mAP. Next, I trained 100 epochs using Pascal VOC, and the mAP reached 81%, which is higher than most algorithms.

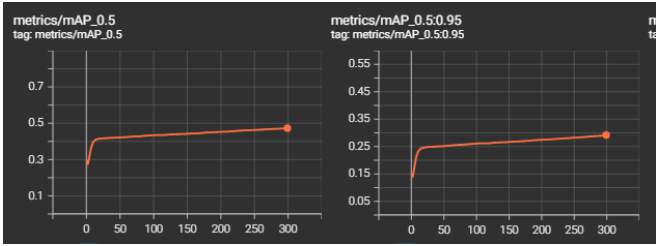

Figure 13: Training 300 epochs on the COCO dataset, mAP steadily increases to 47.4 and shows no signs of convergence

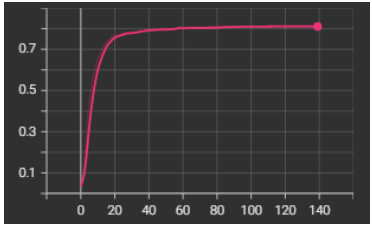

Figure 14: Convergence after training 100 epochs on Pascal VOC dataset, mAP reaches 81%.

Compared with several models using VGG and ResNet as the backbone network, my model has significantly reduced the number of parameters because my model is only 3m, which is very light, while the accuracy has mostly improved. Compared with the YOLOv5n before the improvement, the accuracy is not greatly improved, but the mAP images of the training results still do not show any saturation. If we can retrain more rounds, we will definitely surpass YOLOv5n by more.

| Model | Params(M) | mAP@0.5 | mAP@[0.5, 0.9] |
|---|---|---|---|
| YOLOv5nt | 3 | 47.4 | 29.4 |
| Faster RCNN(vgg16) | 138(-135) | 41.5(+5.9) | 21.2(+7.8) |
| Faster RCNN(resnet101) | 85(-82) | 48.4(-1) | 27.2(+2.2) |
| YOLOv5n | 1.9(+1.1) | 45.7(+1.4) | 28(+1.4) |
| TTFNet-DarkNet | 43(-40) | 50(-2.6) | -- |
| RetinaNet(ResNet18) | 33(-30) | 44.9(+2.5) | -- |

Table 1: Comparison of training results on COCO dataset with other object detection models

| Model | Params(M) | mAP |
|---|---|---|
| YOLOv5nt | 3 | 81 |
| faster RCNN(vgg16) | 138(-135) | 73.2(+7.8) |
| faster RCNN(resnet101) | 85(-82) | 76.4(+4.6) |
| SSDlite(mobilenetv3) | 3.82(-0.82) | 69.2(+11.8) |
| FSSD(VGG16) | 138(-135) | 80.4(+0.6) |

Table 2: Comparison of training results on Pascal VOC dataset with other object detection models

I think one of the reasons why my model does not saturate is because of the features of the transformer. In the vision transformer experiments, we can see that the characteristics of the transformer compared to the CNN are that the accuracy rate can exceed that of the CNN only if the pre-training data set is large enough, and at this point the CNN is saturated while the transformer is not saturated. I tried to save and retrain the model after three hundred epochs of training, and the mAP dropped dramatically because the model parameters were inherited, but the learning rate and other parameters were not inherited, so the loss increased dramatically after iteration. After I checked the official GitHub of yolov5, the solution given by the author "about unsaturated after 300 rounds of training" is: train 500 rounds, 1000 rounds, etc. from the beginning. I did not retrain because it was too time consuming.

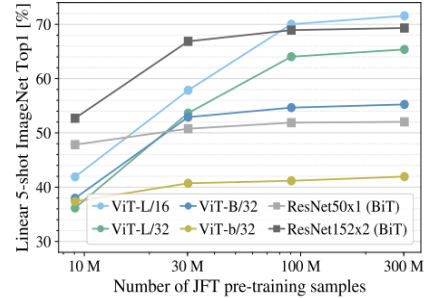

Figure 15: Comparison of CNN and vision transformer results after pre-training on different size datasets

However, the model still has the shortcoming that the transformer part in the conv-transformer module, which is designed to speed up the training, is too simple due to the lack of time. In the vision transformer, the calculation of multi-headed self-attention uses 12 heads, but in my model only 4 heads are used to simplify the calculation. While the vision transformer has 12 repetitions of self-attention stacking, my model has only 1. This makes my model lacking in attention feature extraction.

## VI. CONCLUSION

In this paper, a convolutional module with transformer is proposed to fuse the detailed features extracted by CNN with the global features extracted by transformer, in order to improve the recognition accuracy of the model. The main execution steps are: convolutional downsampling, reduction of feature map size, self-attention calculation and up-sampling, and finally stitching and fusion with the initial input. In the experimental part, the module was applied to the end of YOLOv5n, and after training 300 epochs on the coco dataset, the map was improved by 1.7% compared to the previous YOLOv5n, and the map curve did not show saturation, so there is still potential for improvement. After 100 rounds of training on the pascal voc dataset, the result achieves 81% accuracy, which is 4.6 percent better than the faster RCNN using resnet101 as the backbone network, but the number of parameters is less than one-twentieth of it.

## VII. FUTURE WORK

First of all, the biggest problem is that the coco dataset is too large, so the model is too simple to build in order to speed up the training, for example, the improvement is based on YOLOv5n, its model depth and channel depth are not enough, and the conv-transformer module self-attention only uses 4 heads and only repeats once. The simplified model cannot extract enough features. In the future, the model can be deepened and the accuracy can definitely be improved significantly (refer to YOLOv5n and YOLOv5l). The second is for the improvement of the model is only added to the end of the yolov5n, later I can try to use the conv-transformer to build more structures, such as replacing the SPPF module or put into the PAN architecture. In addition, because of the lack of time and equipment, I could not train more and finally failed to converge the loss function. In the future, if better equipment is available, such as Tesla v100, it will be possible to train until saturation.

Future work will focus on how to fuse the global features of the transformer with the CNN and reduce the computational cost of the transformer. In the swin-transformer[21], the authors propose a structure that locally computes self-attention and continuously downsamples. I think this structure can be completely replaced by convolutional operations, and I am going to study a parallel structure of transformer and CNN.

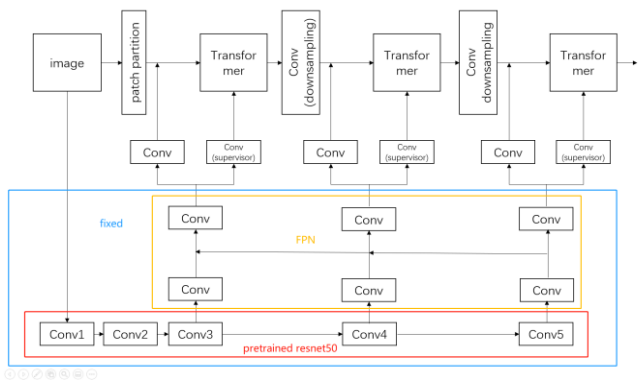

Figure 16: My previously proposed transformer-CNN parallel structure, which could not be implemented due to time

# MSc Project - Reflective Essay

| Project Title: | CNN-transformer mixed model for object detection |
|---|---|
| Student Name: | Wenshuo Li |
| Student Number: | 210276415 |
| Supervisor Name: | Ioannis Patras |
| Programme of Study: | Machine Learning for Visual Data Analytics |

1. Strengths/Weaknesses

1.1 Strengths

The main advantage of the proposed model in this paper is the small amount of parameters, compared to the general model with several tens of m parameters, the number of parameters in my model is only 3m.

In this paper, the proposed conv-transformer module operates by convolutional downsampling, decreasing the feature map size, then self-attention[1] calculation and upsampling, and finally splicing and fusion with the initial input. It can fuse the detailed features extracted by CNN[2] with the global features extracted by transformer[3] to improve the recognition accuracy of the model. The computation of the transformer module is significantly reduced by convolutional downsampling and upsampling.

After training the model on coco dataset with 300 epochs, the accuracy reached 47.4, which is 1.7% higher than the yolov5n[4] before improvement, but the map curve did not converge and was still increasing, if we retrain more epochs, we will get higher mAP.

I think one of the reasons why my model does not saturate is because of the features of the transformer. In the vision transformer experiments, we can see that the characteristics of the transformer compared to the CNN are that the accuracy rate can exceed that of the CNN only if the pre-training data set is large enough, and at this point the CNN is saturated while the transformer is not saturated.

1.2 Weaknesses

However, the model still has the shortcoming that the transformer part in the conv-transformer module, which is designed to speed up the training, is too simple due to the lack of time. In the vision transformer, the calculation of multi-headed self-attention uses 12 heads, but in my model only 4 heads are used to simplify the calculation. While the vision transformer has 12 repetitions of self-attention stacking, my model has only 1. This makes my model lacking in attention feature extraction.

2. Presentation of possibilities for further work

First of all, the biggest problem is that the coco dataset is too large, so the model is too simple to build in order to speed up the training, for example, the improvement is based on YOLOv5n, its model depth and channel depth are not enough, and the conv-transformer module self-attention only uses 4 heads and only repeats once. The simplified model cannot extract enough features. In the future, the model can be deepened and the accuracy can definitely be improved significantly (refer to YOLOv5n and YOLOv5l). The second is for the improvement of the model is only added to the end of the yolov5n, later I can try to use the conv-transformer to build more structures, such as replacing the SPPF[5] module or put into the PAN[6] architecture. In addition, because of the lack of time and equipment, I could not train more and finally failed to converge the loss function. In the future, if better equipment is available, such as Tesla v100, it will be possible to train until saturation.

Future work will focus on how to fuse the global features of the transformer with the CNN and reduce the computational cost of the transformer. In the swin-transformer[7], the authors propose a structure that locally computes self-attention and continuously downsamples. I think this structure can be completely replaced by convolutional operations, and I am going to study a parallel structure of transformer and CNN.

3.The whole project process and reflection

In the process of completing this project I learned a lot about how difficult it is to conduct a research project. When I first decided on the topic, I only knew that it was a related object detection project, and after asking other students, I found that what everyone wanted to do was to use existing methods to apply to their own datasets, such as using YOLO to detect specific buildings on satellite maps, or to detect enemies in computer games.I personally think that this kind of application project will be very easy to complete, because you only need to label the location and category of the object to be detected in the image, and the final detection accuracy will be very high. This is because the dataset that you label yourself is usually very small, in the case of a small dataset, the model is easy to fit. For example, I have used coco128 for model pre-training, the training set is only the first 128 images in the coco dataset, the training results quickly reached an amazing accuracy of more than 95 percent.

With my confusion I started to read a lot of papers related to object detection, which is a very laborious process, because each paper is the result of a large number of papers built on previous ones. In order to understand the idea of a paper, I had to go back a long time and start reading the papers in order. Many papers must be read several times and combined with the code provided in the paper to understand the exact operation. In this process I gradually learned how the models related to target detection were improved step by step. The ingenuity of the methods proposed by these authors gave me a sense of the greatness of these papers. The best papers on computer vision basically improve the previous model so that the model can extract features that are more suitable for target detection. So, I planned to improve the model as well, but the hardest part of this topic is how to come up with a way to improve the model. I spent a lot of time reading papers to understand the implications of their model improvements and the advantages and disadvantages of various models. I finally settled on a hybrid model using convolutional neural network and transformer.

I was going to use transformer for backbone network at first, but considering that the computation of transformer is too large and it needs a large dataset to train enough epochs to outperform cnn, I finally gave up the transformer model. Because of the time spent on reading the paper and the lack of computing speed of the device, it was impossible to train a model with a transfomer as the backbone network. So I came up with the idea of using a fast CNN as the backbone network and incorporating the transformer module at the end.

After many days of conceptualizing and finally deciding on the conv-transformer module, I then started writing the code. The code was chosen from the pytorch framework because of the clarity of the pytorch framework for building modules. After building the model I started testing on coco128 using my own card rtx 2060, after a few dimensional issues and tweaking it to train normally, when I found the coco dataset had about 110k images I realized I couldn't finish before the deadline.So I used Google Colab pro+ to rent a Tesla p100 for training. the advantage of Google colab is that you can use your phone or tablet to open the web page, so you can train the model 24 hours. After the coco dataset was trained, the results exceeded the previous model, but not saturated, there is still room for improvement, but there is no time to train from the beginning. So I looked at the dataset commonly used in papers in target detection has a relatively small dataset pascal voc, I used the model trained on coco on pascal voc to do migration learning and achieve good results.

4.Possible legal ethical issues involved in the project

I think the most likely law to be violated by target detection tasks is invasion of privacy. This is because target detection generally uses cameras to acquire video for real-time detection. With the target detection method in the camera, we can get data about the characteristics and direction of movement of the person in the current camera, and people do not know what the owner of the camera will do about using this data. For example, some people may use the video data captured by street cameras every day to get the travel routine of the people living around them, which is quite scary.

I think everyone in the vicinity of each camera has the right to know what the data captured by that camera is being used for. To protect everyone's privacy, I think the law should require that signs be placed under each camera telling people who owns the camera and what the data is being used for.For example, the camera belongs to the reception of the apartment and its role is to protect the security of the residents of the apartment, or the camera belongs to the shopping mall and the data obtained is used to analyze purchase records for restocking purposes. If this is done it will ensure that people's data is not used for illegal purposes.